\documentclass{article}
\usepackage{amssymb}
\usepackage{amsmath}
\usepackage{microtype}
\usepackage{graphicx}
\usepackage{booktabs}
\usepackage{subcaption}
\usepackage{hyperref}
\usepackage[capitalize,nameinlink,noabbrev]{cleveref}
\usepackage[autostyle]{csquotes}
\usepackage{enumitem}
\usepackage{multirow}
\usepackage{times}
\usepackage{textcomp}
\usepackage[T1]{fontenc}

\makeatletter
\newlength\mylen
\DeclareRobustCommand*\Textsuperscript[1]{%
\@Textsuperscript{\selectfont#1}}
\def\@Textsuperscript#1{%
\settoheight\mylen{\fontsize\f@size\z@ A}%
{\m@th\ensuremath{\raise.3\mylen\hbox{\fontsize\sf@size\z@#1}}}}

\makeatother

\usepackage[accepted]{icml2021}

\icmltitlerunning{ViLT: Vision-and-Language Transformer Without Convolution or Region Supervision}

\begin{document}

\twocolumn[
\icmltitle{ViLT: Vision-and-Language Transformer\\
           Without Convolution or Region Supervision}

\icmlsetsymbol{equal}{*}
\icmlsetsymbol{na}{\textdagger}

\begin{icmlauthorlist}
\icmlauthor{Wonjae Kim}{equal,ke,na}
\icmlauthor{Bokyung Son}{equal,ke}
\icmlauthor{Ildoo Kim}{kb}
\end{icmlauthorlist}

\icmlaffiliation{ke}{Kakao Enterprise, Seongnam, Gyeonggi, Republic of Korea}
\icmlaffiliation{kb}{Kakao Brain, Seongnam, Gyeonggi, Republic of Korea}

\icmlcorrespondingauthor{Wonjae Kim}{wonjae.kim@navercorp.com}

\icmlkeywords{Machine Learning, ICML, Vision-and-Language Pre-training, Visual Question Answering, Image Retrieval, Text Retrieval, Visual Reasoning, Image Captioning, Transformers, Attention Mechanisms, Vision Transformers}

\vskip 0.3in
]
\printAffiliationsAndNotice{\icmlEqualContribution \textsuperscript{\textdagger}Current affiliation: NAVER AI Lab, Seongnam, Gyeonggi, Republic of Korea.} % otherwise use the standard text.

\begin{abstract}
    Vision-and-Language Pre-training (VLP) has improved performance on various joint vision-and-language downstream tasks.
    Current approaches to VLP heavily rely on image feature extraction processes, most of which involve region supervision (e.g., object detection) and the convolutional architecture (e.g., ResNet).
    Although disregarded in the literature, we find it problematic in terms of both (1) efficiency/speed, that simply extracting input features requires much more computation than the multimodal interaction steps; and (2) expressive power, as it is upper bounded to the expressive power of the visual embedder and its predefined visual vocabulary.
    In this paper, we present a minimal VLP model, Vision-and-Language Transformer (ViLT), monolithic in the sense that the processing of visual inputs is drastically simplified to just the same convolution-free manner that we process textual inputs.
    We show that ViLT is up to tens of times faster than previous VLP models, yet with competitive or better downstream task performance.
    Our code and pre-trained weights are available at \url{https://github.com/dandelin/vilt}.
\end{abstract}

\begin{figure}[t]
    \centering
    \includegraphics[width=0.95\linewidth]{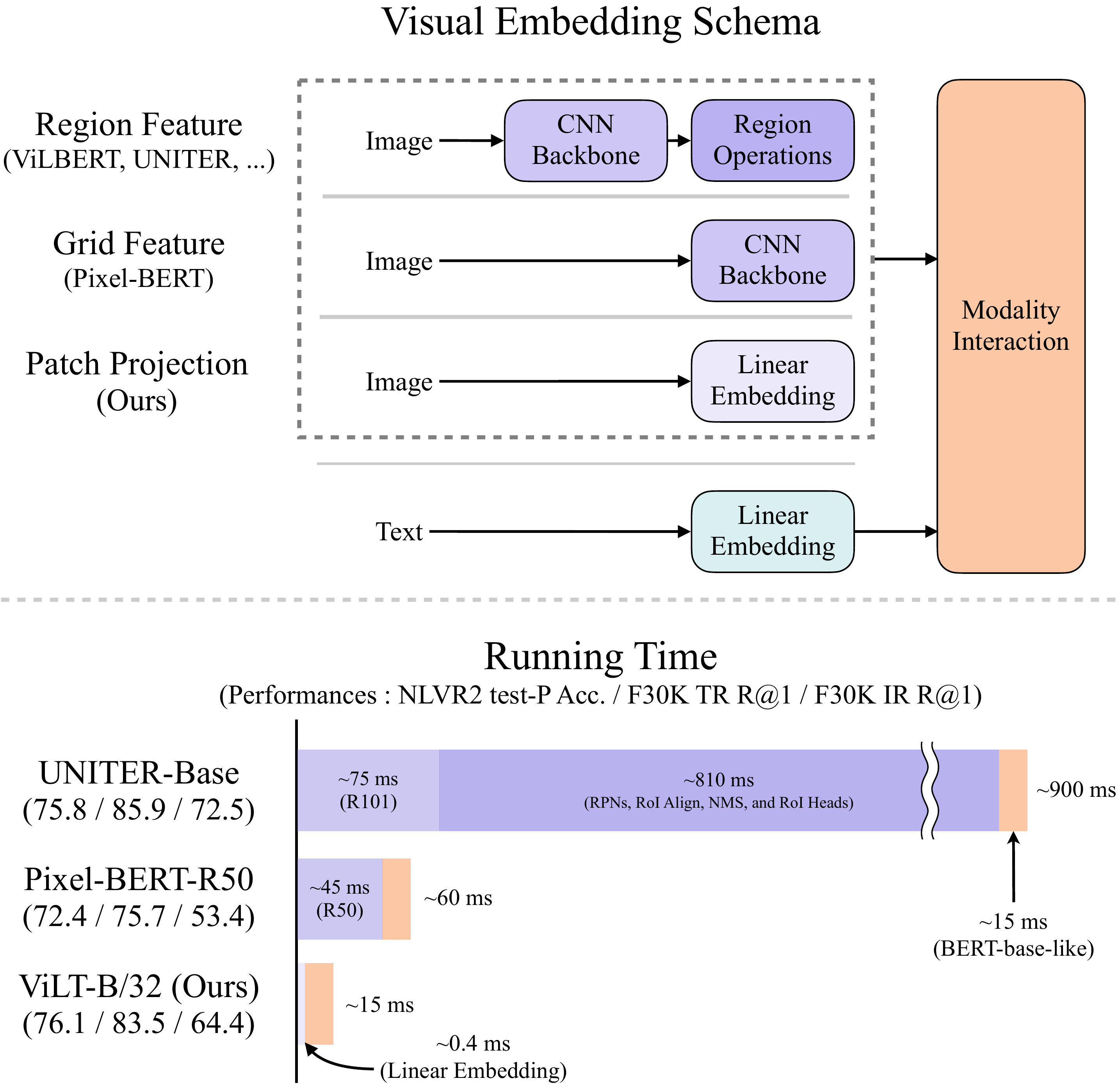}
    \caption{
        Visual comparison of conventional VLP architectures and our proposed ViLT.
        We have entirely removed convolutional neural networks from the VLP pipeline without hurting performance on downstream tasks.
        ViLT is the first VLP model of which the modal-specific components require \textit{less} computation than the transformer component for multimodal interactions.
    }
    \label{fig:pipeline}
\end{figure}

\begin{figure*}[t]
    \begin{center}
    \begin{subfigure}[b]{.25\textwidth}
        \centering
        \includegraphics[width=0.45\linewidth]{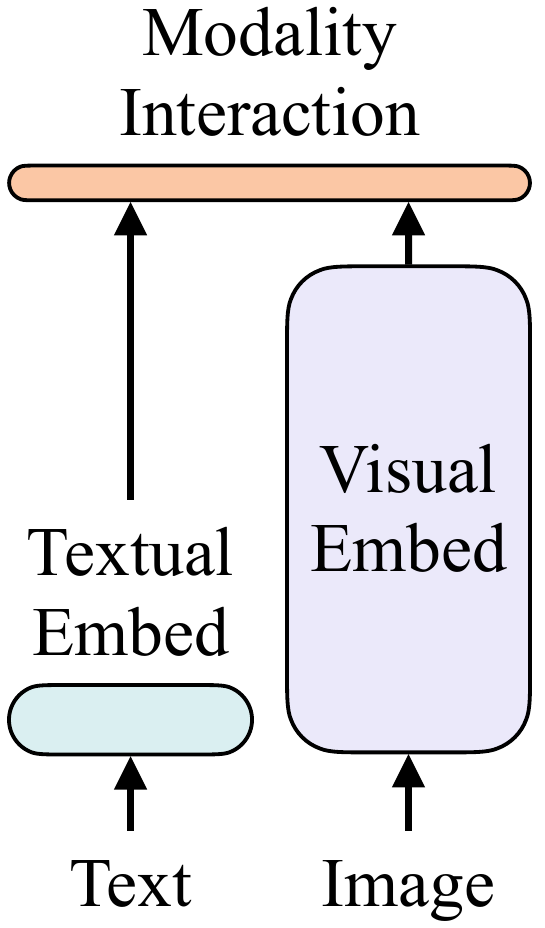}
        \caption{$ \text{VE} > \text{TE} > \text{MI} $}
        \label{sfig:uneven_disentangled}
    \end{subfigure}\hfill
    \begin{subfigure}[b]{.25\textwidth}
        \centering
        \includegraphics[width=0.45\linewidth]{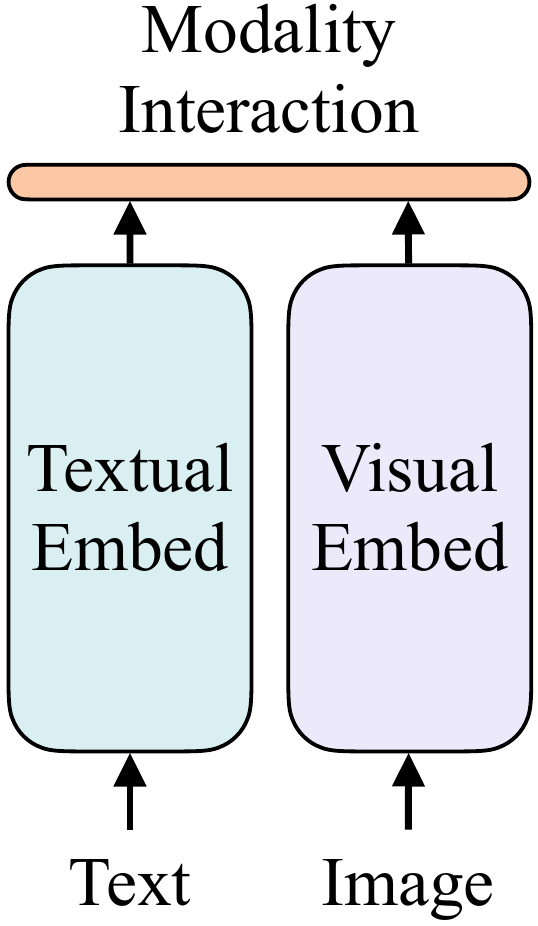}
        \caption{$ \text{VE} = \text{TE} > \text{MI} $}
        \label{sfig:even_disentangled}
    \end{subfigure}\hfill
    \begin{subfigure}[b]{.25\textwidth}
        \centering
        \includegraphics[width=0.45\linewidth]{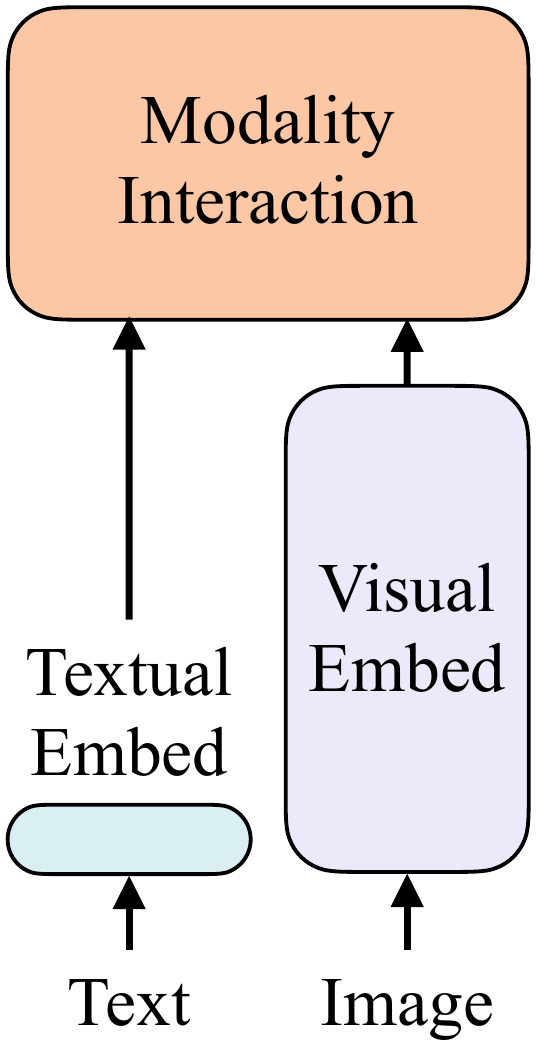}
        \caption{$ \text{VE} > \text{MI} > \text{TE} $}
        \label{sfig:uneven_entangled}
    \end{subfigure}\hfill
    \begin{subfigure}[b]{.25\textwidth}
        \centering
        \includegraphics[width=0.45\linewidth]{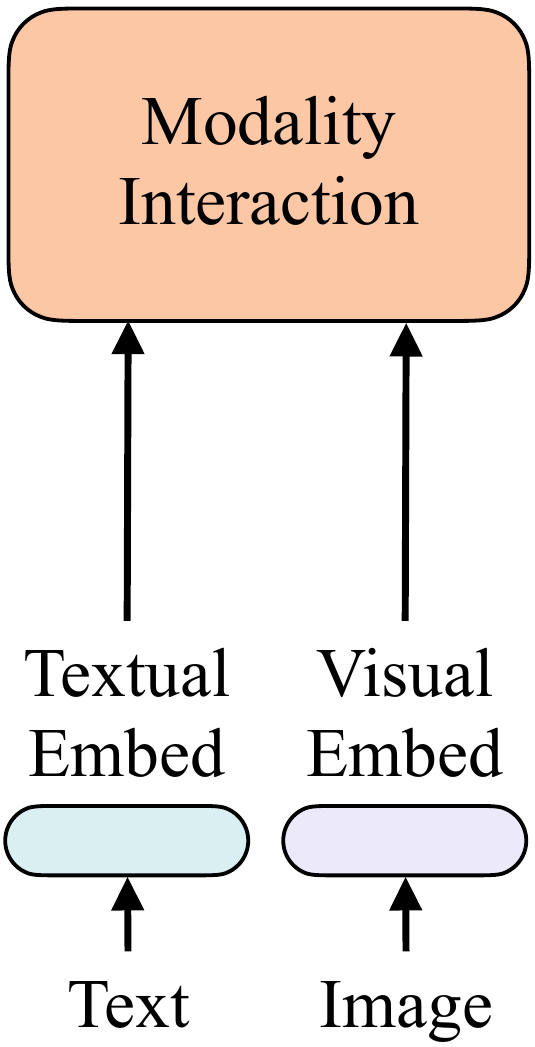}
        \caption{$ \text{MI} > \text{VE} = \text{TE} $}
        \label{sfig:even_entangled}
    \end{subfigure}\hfill
    \caption{Four categories of vision-and-language models. The height of each rectangle denotes its relative computational size. VE, TE, and MI are short for visual embedder, textual embedder, and modality interaction, respectively.}
    \label{fig:taxonomy}
    \end{center}
\end{figure*}

\section{Introduction}
\label{sec:intro}

The pre-train-and-fine-tune scheme has been expanded to a joint domain of vision and language, giving birth to the category of \textit{Vision-and-Language Pre-training (VLP)} models \citep{lu2019vilbert, chen2019uniter,su2019vl,li2019visualbert,tan2019lxmert,li2020unicoder,lu202012,cho2020x,qi2020imagebert,zhou2020unified,huang2020pixel,li2020oscar,gan2020large,yu2020ernie,zhang2021vinvl}.
These models are pre-trained with image text matching and masked language modeling objectives\footnote{While some works employ additional objectives and data structures, these two objectives apply to almost every VLP model.} on images and their aligned descriptions, and are fine-tuned on vision-and-language downstream tasks where the inputs involve two modalities. 

To be fed into VLP models, image pixels need to be initially embedded in a dense form alongside language tokens.
Since the seminal work of \citet{krizhevsky2012imagenet}, deep convolutional networks have been regarded as essential for this visual embedding step. 
Most VLP models employ an object detector pre-trained on the Visual Genome dataset \citep{krishna2017visual} annotated with 1,600 object classes and 400 attribute classes as in \citet{anderson2018bottom}.
Pixel-BERT \citep{huang2020pixel} is one exception of this trend, as it uses ResNet variants \citep{he2016deep, xie2017aggregated} pre-trained on ImageNet classification \citep{russakovsky2015imagenet} embedding pixels in place of object detection modules.

To this date, most VLP studies have focused on improving performance by increasing the power of visual embedders.
The shortcomings of having a heavy visual embedder are often disregarded in academic experiments because region features are commonly cached in advance at training time to ease the burden of feature extraction. However, the limitations are still evident in real-world applications as the queries in the wild have to undergo a slow extraction process.

To this end, we shift our attention to the lightweight and fast embedding of visual inputs.
Recent work \citep{dosovitskiy2020image, touvron2020training} demonstrated that using a simple linear projection of a patch is effective enough to embed pixels before feeding them into transformers.
Whereas being the solid mainstream for text \citep{devlin2019bert}, it is only recently that transformers \citep{vaswani2017attention} are used for images as well.
We presume that the transformer module-- used for modality interaction in VLP models-- can also manage to process visual features in place of a convolutional visual embedder, just as it processes textual features.

This paper proposes the Vision-and-Language Transformer (ViLT) that handles two modalities in a single unified manner.
It mainly differs from previous VLP models in its shallow, convolution-free embedding of pixel-level inputs.
Removing deep embedders solely dedicated to visual inputs significantly cuts down the model size and running time by design.
\cref{fig:pipeline} shows that our parameter-efficient model is tens of times faster than VLP models with region features and at least four times faster than those with grid features while exhibiting similar or even better performance on vision-and-language downstream tasks.

Our key contributions can be summarized as follows:
\begin{itemize}
    \item ViLT is the \textit{simplest} architecture by far for a vision-and-language model as it commissions the transformer module to extract and process visual features in place of a separate deep visual embedder. This design inherently leads to significant runtime and parameter efficiency.
    \item For the first time, we achieve competent performance on vision-and-language tasks without using region features or deep convolutional visual embedders in general. 
    \item Also, for the first time, we empirically show that whole word masking and image augmentations that were unprecedented in VLP training schemes further drive downstream performance.
\end{itemize}

\section{Background}
\label{sec:background}

\subsection{Taxonomy of Vision-and-Language Models}

We propose a taxonomy of vision-and-language models based on two points: (1) whether the two modalities have an even level of expressiveness in terms of dedicated parameters and/or computation; and (2) whether the two modalities interact in a deep network.
A combination of these points leads to four archetypes in \cref{fig:taxonomy}.

The \textit{visual semantic embedding} (VSE) models such as VSE++ \citep{faghri2017vse++} and SCAN \citep{lee2018stacked} belong to \cref{sfig:uneven_disentangled}.
They use separate embedders for image and text, with the former being much heavier.
Then, they represent the similarity of the embedded features from the two modalities with simple dot products or shallow attention layers.

CLIP \citep{radford2021learning} belongs to \cref{sfig:even_disentangled} as it uses separate but equally expensive transformer embedders for each modality.
Interaction between the pooled image vector and text vector is still shallow (dot product).
Despite CLIP's remarkable zero-shot performance on image-to-text retrieval, we could not observe the same level of performance on other vision-and-language downstream tasks.
For instance, fine-tuning the MLP head on NLVR2 \citep{suhr2018corpus} with the dot product of pooled visual and textual vectors from CLIP as the multimodal representation gives a low dev accuracy of 50.99 $\pm$ 0.38
(ran with three different seeds);
as chance level accuracy is 0.5, we conclude that the representations are incapable of learning this task.
It also matches the findings of \citet{suhr2018corpus} that all models with simply fused multimodal representation failed to learn NLVR2.

This result backs up our speculation that simple fusion of outputs even from high-performing unimodal embedders may not be sufficient to learn complex vision-and-language tasks, bolstering the need for a more rigorous inter-modal interaction scheme.

Unlike models with shallow interaction, the more recent VLP models that fall under \cref{sfig:uneven_entangled} use a deep transformer to model the interaction of image and text features.
Aside from the interaction module, however, convolutional networks are still involved in extracting and embedding image features, which accounts for most of the computation as depicted in \cref{fig:pipeline}.
Modulation-based vision-and-language models \citep{perez2018film,nguyen2020revisiting} also fall under \cref{sfig:uneven_entangled}, with their visual CNN stems corresponding to visual embedder, RNNs producing the modulation parameters to textual embedder, and modulated CNNs to modality interaction.

Our proposed ViLT is the first model of type \cref{sfig:even_entangled} where the embedding layers of raw pixels are shallow and computationally light as of text tokens.
This architecture thereby concentrates most of the computation on modeling modality interactions.

\subsection{Modality Interaction Schema}

At the very core of contemporary VLP models lie transformers.
They get visual and textual embedding sequences as input, model inter-modal and optionally intra-modal interactions throughout layers, then output a contextualized feature sequence. 

\citet{bugliarello2020multimodal} classifies interaction schema into two categories: (1) \textit{single-stream} approaches (e.g., VisualBERT \citep{li2019visualbert}, UNITER \citep{chen2019uniter}) where layers collectively operate on a concatenation of image and text inputs; and (2) \textit{dual-stream} approaches (e.g., ViLBERT \citep{lu2019vilbert}, LXMERT \citep{tan2019lxmert}) where the two modalities are not concatenated at the input level. 
We follow the single-stream approach for our interaction transformer module because the dual-stream approach introduces additional parameters.

\begin{figure*}[t]
    \centering
    \includegraphics[width=0.95\textwidth]{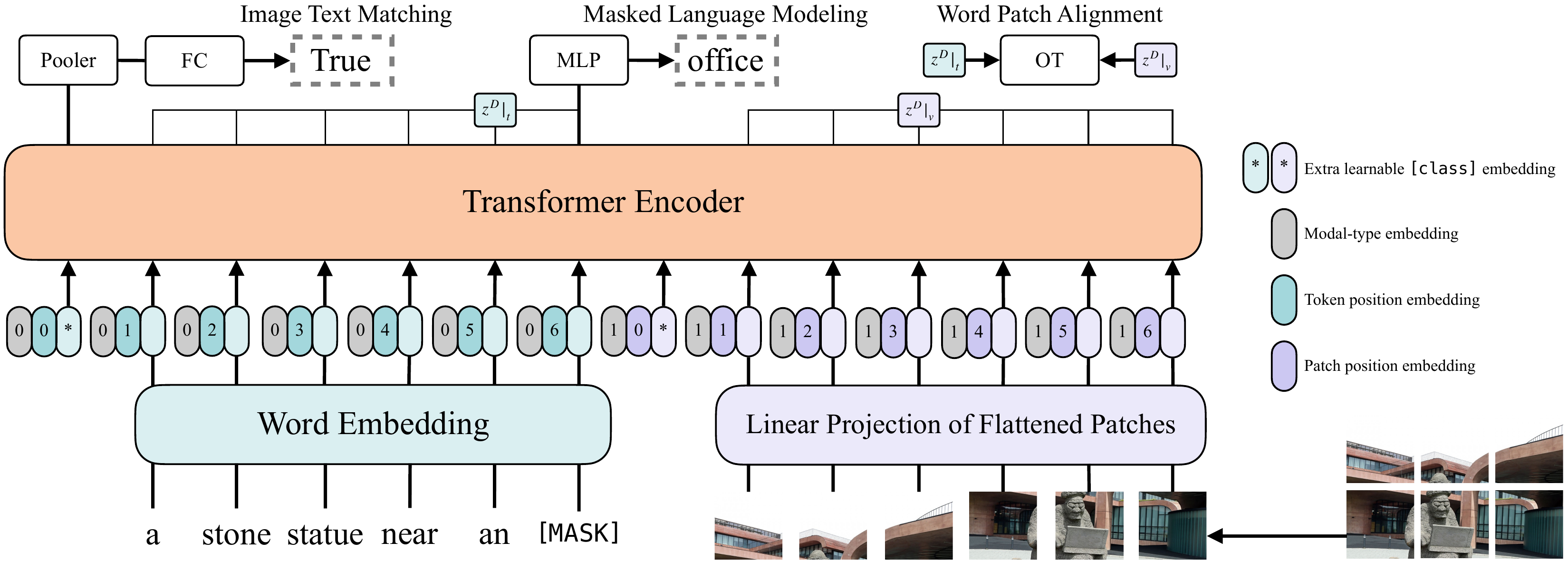}
    \caption{Model overview. Illustration inspired by \citet{dosovitskiy2020image}.}
    \label{fig:vilt}
\end{figure*}

\subsection{Visual Embedding Schema}

Whereas all performant VLP models share the same textual embedder-- tokenizer from pre-trained BERT, word and position embeddings resembling those of BERT-- they differ on visual embedders.
Still, in most (if not all) cases, visual embedding is the bottleneck of existing VLP models.
We focus on cutting corners on this step by introducing patch projection instead of using region or grid features for which heavy extraction modules are used.

\paragraph{Region Feature.}
VLP models dominantly utilize region features, also known as bottom-up features \citep{anderson2018bottom}.
They are obtained from an off-the-shelf object detector like Faster R-CNN \citep{ren2016faster}.

The general pipeline of generating region features is as follows. First, a region proposal network (RPN) proposes regions of interest (RoI) based on the grid features pooled from the CNN backbone. Non-maximum suppression (NMS) then reduces the number of RoIs to a few thousand. After being pooled by operations such as RoI Align \citep{he2017mask}, the RoIs go through RoI heads and become region features. NMS is again applied to every class, finally reducing the number of features under a hundred.

The above process involves several factors that affect the performance and runtime: the backbone, the style of NMS, the RoI heads. Previous works were lenient with controlling these factors, making varying choices from each other as listed in \cref{tab:vlp_components}.\footnote{\citet{bugliarello2020multimodal} showed that a controlled setup bridges the performance gap of various region-feature-based VLP models.}

\begin{itemize}
    \item Backbone: ResNet-101 \citep{lu2019vilbert,tan2019lxmert,su2019vl} and ResNext-152 \citep{li2019visualbert,li2020unicoder,zhang2021vinvl} are two commonly used backbones.
    \item NMS: NMS is typically done in a \textit{per-class} fashion. Applying NMS to each and every class becomes a major runtime bottleneck with a large number of classes, \textit{e.g.} 1.6K in the VG dataset \citep{jiang2020defense}. \textit{Class-agnostic} NMS was recently introduced to tackle this issue \citep{zhang2021vinvl}.
    \item RoI head: C4 heads were initially used \citep{anderson2018bottom}. FPN-MLP heads were introduced later \citep{jiang2018pythia}. As heads operate for each and every RoI, they pose a substantial runtime burden.
\end{itemize}

However lightweight, object detectors are less likely to be faster than the backbone or a single-layer convolution. Freezing the visual backbone and caching the region features in advance only helps at training time and not during inference, not to mention that it could hold performance back.

\paragraph{Grid Feature.}
Besides detector heads, the output feature grid of convolutional neural networks such as ResNets can also be used as visual features for vision-and-language pre-training.
Direct use of grid features was first proposed by VQA-specific models \citep{jiang2020defense, nguyen2020revisiting}, mainly to avoid using severely slow region selection operations.

X-LXMERT \citep{cho2020x} revisited grid features by fixing the region proposals to grids instead of those from the region proposal networks. However, their caching of features excluded further tuning of the backbone.

Pixel-BERT is the only VLP model that replaces the VG-pre-trained object detector with a ResNet variant backbone pre-trained with ImageNet classification. Unlike frozen detectors in region-feature-based VLP models, the backbone of Pixel-BERT is tuned during vision-and-language pre-training.
The downstream performance of Pixel-BERT with ResNet-50 falls below region-feature-based VLP models, but it matches that of other competitors with the use of a much heavier ResNeXt-152.

We claim that grid features are not the go-to option, however, since deep CNNs are still expensive that they account for a large portion of the whole computation as in \cref{fig:pipeline}.

\paragraph{Patch Projection.}
To minimize overhead, we adopt the simplest visual embedding scheme: \textit{linear projection} that operates on image patches.
The patch projection embedding was introduced by ViT \citep{dosovitskiy2020image} for image classification tasks.
Patch projection drastically simplifies the visual embedding step to the level of textual embedding, which also consists of simple projection (lookup) operations.

We use a 32 $\times$ 32 patch projection which only requires 2.4M parameters. This is in sharp contrast to complex ResNe(X)t backbones\footnote{Parameters for R50 is 25M, R101 is 44M, and X152 is 60M.} and detection components.
Its running time is also ignorable as shown in \cref{fig:pipeline}.
We make a detailed runtime analysis in \cref{subsec:runtime}.

\section{Vision-and-Language Transformer}
\label{sec:vilt}

\subsection{Model Overview}
ViLT has a succinct architecture as a VLP model with a minimal visual embedding pipeline and following the single-stream approach.

We deviate from the literature that we initialize the interaction transformer weights from pre-trained ViT instead of BERT.
Such initialization exploits the power of the interaction layers to process visual features while lacking a separate deep visual embedder.
\footnote{We also experimented with initializing the layers from BERT weights and using the pre-trained patch projection from ViT, but it did not work.}
\begin{align} 
    \bar{t} &= [t_{\text{class}}; t_1 T; \cdots; t_L T] + T^{\text{pos}} \\
    \bar{v} &= [v_{\text{class}}; v_1 V; \cdots; v_N V] + V^{\text{pos}} \\
    z^0 &= [\bar{t} + t^{\text{type}}; \bar{v} + v^{\text{type}}] \\
    \hat{z}^{d} &= \text{MSA}(\text{LN}(z^{d-1})) + z^{d-1}, &d = 1 \ldots D \\
    z^d &= \text{MLP}(\text{LN}(\hat{z}^{d})) + \hat{z}^{d}, &d = 1 \ldots D \\
    p &= \text{tanh}(z^D_0 W_{\text{pool}})
\end{align}
ViT consists of stacked blocks that include a multiheaded self-attention (MSA) layer and an MLP layer. The position of layer normalization (LN) in ViT is the only difference from BERT: LN comes after MSA and MLP in BERT (``post-norm'') and before in ViT (``pre-norm'').
The input text $t \in \mathbb{R}^{L \times |V|}$ is embedded to $\bar{t} \in \mathbb{R}^{L \times H}$ with a word embedding matrix $T \in \mathbb{R}^{|V| \times H}$ and a position embedding matrix $T^{\text{pos}} \in \mathbb{R}^{(L+1) \times H}$.

The input image $I \in \mathbb{R}^{C \times H \times W}$ is sliced into patches and flattened to $v \in \mathbb{R}^{N \times (P^2 \cdot C)}$ where $(P, P)$ is the patch resolution and $N = HW/P^2$. Followed by linear projection $V \in \mathbb{R}^{(P^2 \cdot C) \times H}$ and position embedding $V^{\text{pos}} \in \mathbb{R}^{(N+1) \times H}$, $v$ is embedded into $\bar{v} \in \mathbb{R}^{N \times H}$.

The text and image embeddings are summed with their corresponding modal-type embedding vectors $t^{\text{type}}, v^{\text{type}} \in \mathbb{R}^{H}$, then are concatenated into a combined sequence $z^0$. The contextualized vector $z$ is iteratively updated through $D$-depth transformer layers up until the final contextualized sequence $z^D$. $p$ is a pooled representation of the whole multimodal input, and is obtained by applying linear projection $W_{\text{pool}} \in \mathbb{R}^{H \times H}$ and hyperbolic tangent upon the first index of sequence $z^D$.

For all experiments, we use weights from ViT-B/32 pre-trained on ImageNet, hence the name ViLT-B/32.\footnote{ViT-B/32 is pre-trained with ImageNet-21K and fine-tuned on ImageNet-1K for image classification. We expect that weights pre-trained on larger datasets (e.g., JFT-300M) would yield better performance.}
Hidden size $H$ is 768, layer depth $D$ is 12, patch size $P$ is 32, MLP size is 3,072, and the number of attention heads is 12.

\subsection{Pre-training Objectives}
We train ViLT with two objectives commonly used to train VLP models: image text matching (ITM) and masked language modeling (MLM).

\paragraph{Image Text Matching.}
We randomly replace the aligned image with a different image with the probability of 0.5.
A single linear layer ITM head projects the pooled output feature $p$ to logits over binary class, and we compute negative log-likelihood loss as our ITM loss.

Plus, inspired by the word region alignment objective in \citet{chen2019uniter}, we design word patch alignment (WPA) that computes the alignment score between two subsets of $z^D$: $z^D|_t$ (textual subset) and $z^D|_v$ (visual subset), using the inexact proximal point method for optimal transports (IPOT) \citep{xie2020fast}.
We set the hyperparameters of IPOT following \citet{chen2019uniter} ($\beta = 0.5, N=50$), and add the approximate wasserstein distance multiplied by 0.1 to the ITM loss.

\paragraph{Masked Language Modeling.}
This objective is to predict the ground truth labels of masked text tokens $t_{\text{masked}}$ from its contextualized vector $z^D_{\text{masked}}|_t$.
Following the heuristics of \citet{devlin2019bert}, we randomly mask $t$ with the probability of 0.15.

We use a two-layer MLP MLM head that inputs $z^D_{\text{masked}}|_t$ and outputs logits over vocabulary, just as the MLM objective of BERT.
The MLM loss is then computed as the negative log-likelihood loss for the masked tokens.

\subsection{Whole Word Masking}
\label{sec:wwm}
Whole word masking is a masking technique that masks all consecutive subword tokens that compose a whole word.
It is shown to be effective on downstream tasks when applied to original and Chinese BERT \citep{cui2019pre}.

We hypothesize that whole word masking is particularly crucial for VLP in order to make full use of information from the other modality.
For example, the word \enquote{giraffe} is tokenized into three wordpiece tokens \texttt{\lbrack"gi", "\#\#raf", "\#\#fe"\rbrack} with the pre-trained \texttt{bert-base-uncased} tokenizer. If not all tokens are masked, say, \texttt{\lbrack"gi", "[MASK]", "\#\#fe"\rbrack}, the model may solely rely on the nearby two language tokens \texttt{\lbrack"gi", "\#\#fe"\rbrack} to predict the masked \texttt{"\#\#raf"} rather than using the information from the image.

We mask whole words with a mask probability of 0.15 during pre-training.
We discuss its impact in \cref{sec:ablation}.

\subsection{Image Augmentation}

Image augmentation reportedly improves the generalization power of vision models \citep{shorten2019survey}.
DeiT \citep{touvron2020training} that builds on ViT experimented with various augmentation techniques \citep{zhang2017mixup,yun2019cutmix,berman2019multigrain,hoffer2020augment,cubuk2020randaugment}, and found them beneficial for ViT training.
However, the effects of image augmentation have not been explored within VLP models.
Caching visual features restrains region-feature-based VLP models from using image augmentation.
Notwithstanding its applicability, neither did Pixel-BERT study its effects.

To this end, we apply RandAugment \citep{cubuk2020randaugment} during fine-tuning.
We use all the original policies except two: color inversion, because texts often contain color information as well, and cutout, as it may clear out small but important objects dispersed throughout the whole image.
We use $N=2, M=9$ as the hyperparameters.
We discuss its impact in \cref{sec:ablation} and \cref{sec:conclusion}.

\section{Experiments}
\label{sec:experiments}

\subsection{Overview}

\begin{table}[t]
    \caption{Pre-training dataset statistics. Caption length is the length of tokens from pre-trained \texttt{bert-base-uncased} tokenizer. {\textdagger} GCC and SBU provide only image urls, so we collect the images from urls which were still accessible.}
    \label{tab:pre-train_datasets}
    \begin{center}
        {\fontsize{10.1pt}{10.1pt}\selectfont
    \begin{tabular}{lccc}
        \toprule
   Dataset & \# Images    & \# Captions  & Caption Length    \\ \midrule
   MSCOCO & 113K    & 567K   & 11.81 $\pm$ 2.81  \\
   VG   & 108K   & 5.41M & 5.53 $\pm$ 1.76   \\
   GCC\Textsuperscript{\textdagger}   & 3.01M & 3.01M & 10.66 $\pm$ 4.93  \\
   SBU\Textsuperscript{\textdagger}  & 867K   & 867K   & 15.0 $\pm$ 7.74  \\ \bottomrule
   \end{tabular}}
\end{center}
\end{table}

We use four datasets for pre-training: Microsoft COCO (MSCOCO) \citep{lin2014microsoft}, Visual Genome (VG) \citep{krishna2017visual}, SBU Captions (SBU) \citep{ordonez2011im2text}, and Google Conceptual Captions (GCC) \citep{sharma2018conceptual}.
\cref{tab:pre-train_datasets} reports the dataset statistics.

We evaluate ViLT on two widely explored types of vision-and-language downstream tasks: for \textit{classification}, we use VQAv2 \citep{goyal2017making} and NLVR2 \citep{suhr2018corpus}, and for \textit{retrieval}, we use MSCOCO and Flickr30K (F30K) \citep{plummer2015flickr30k} re-splited by \citet{karpathy2015deep}.
For the classification tasks, we fine-tune three times with different initialization seeds for the head and data ordering and report the mean scores.
We report the standard deviation in \cref{tab:ablation} along with ablation studies.
For the retrieval tasks, we only fine-tune once.

\subsection{Implementation Details}

For all experiments, we use AdamW optimizer \citep{loshchilov2018decoupled} with base learning rate of $ 10^{-4} $ and weight decay of $ 10^{-2} $.
The learning rate was warmed up for 10\% of the total training steps and was decayed linearly to zero for the rest of the training.
Note that downstream performance may be further improved if we customize the hyperparameters to each task.

We resize the shorter edge of input images to 384 and limit the longer edge to under 640 while preserving the aspect ratio.
This resizing scheme is also used during object detection in other VLP models, but with a larger size of the shorter edge (800).
Patch projection of ViLT-B/32 yields 12 $\times$ 20 = 240 patches for an image with a resolution of 384 $\times$ 640. As this is a rarely reached upper limit, we sample 200 patches at maximum during pre-training.
We interpolate $V^{\text{pos}}$ of ViT-B/32 to fit the size of each image and pad the patches for batch training.
Note that the resulting image resolution is four times smaller than 800 $\times$ 1,333, which is the size that all other VLP models use for inputs to their visual embedders.

We use the \texttt{bert-base-uncased} tokenizer to tokenize text inputs.
Instead of fine-tuning from pre-trained BERT, we learn the textual embedding-related parameters $t_{\text{class}}$, $T$, and $T^{\text{pos}}$ from scratch.
Although beneficial \textit{prima facie}, employing a pre-trained text-only BERT does not guarantee performance gain for vision and language downstream tasks.
Counterevidence has already been reported by \citet{tan2019lxmert}, where initializing with pre-trained BERT parameters led to weaker performance than pre-training from scratch.

We pre-train ViLT-B/32 for 100K or 200K steps on 64 NVIDIA V100 GPUs with a batch size of 4,096.
For all downstream tasks, we train for ten epochs with a batch size of 256 for VQAv2/retrieval tasks and 128 for NLVR2.

\subsection{Classification Tasks}

\begin{table}[t]
    \caption{Comparison of ViLT-B/32 with other models on downstream classification tasks. We use MCAN \citep{yu2019deep} and MaxEnt \cite{suhr2018corpus} for VQAv2 and NLVR2 w/o VLP SOTA results. {\textdagger} additionally used GQA, VQAv2, VG-QA for pre-training. {\textdaggerdbl} made additional use of the Open Images \citep{kuznetsova2020open} dataset. {\textcircled{a}} indicates RandAugment is applied during fine-tuning. {\textcircled{+}} indicates model trained for a longer 200K pre-training steps.}
    \label{tab:clas}
    \begin{center}
        {\fontsize{8.1pt}{8.1pt}\selectfont
    \begin{tabular}{llrccc}
        \toprule
        \multirow{2}{*}{\shortstack[l]{Visual\\Embed}} & \multirow{2}{*}{Model} &  \multirow{2}{*}{\shortstack[r]{Time\\(ms)}} & VQAv2 & \multicolumn{2}{c}{NLVR2} \\
   & & & test-dev & dev & test-P \\ \midrule
   \multirow{7}{*}{Region} & w/o VLP SOTA & \textasciitilde 900 & 70.63 & 54.80 & 53.50 \\
   & ViLBERT & \textasciitilde 920 & 70.55 & - & - \\
   & VisualBERT & \textasciitilde 925 & 70.80 & 67.40 & 67.00 \\
   & LXMERT & \textasciitilde 900 & 72.42 & 74.90 & 74.50 \\
   & UNITER-Base & \textasciitilde 900 & 72.70 & 75.85 & 75.80 \\
   & OSCAR-Base\Textsuperscript{\textdagger} & \textasciitilde 900 & 73.16 & 78.07 & 78.36 \\
   & VinVL-Base\Textsuperscript{\textdagger}\Textsuperscript{\textdaggerdbl} & \textasciitilde 650 & 75.95 & 82.05 & 83.08 \\
   \midrule
   \multirow{2}{*}{Grid} & Pixel-BERT-X152 & \textasciitilde 160 & 74.45 & 76.50 & 77.20 \\
    & Pixel-BERT-R50 & \textasciitilde 60 & 71.35 & 71.70 & 72.40 \\
    \midrule
   \multirow{3}{*}{Linear} & ViLT-B/32 & \textasciitilde 15 & 70.33 & 74.41 & 74.57 \\
    & ViLT-B/32\Textsuperscript{\textcircled{a}} & \textasciitilde 15 & 70.85 & 74.91 & 75.57 \\
    & ViLT-B/32\Textsuperscript{\textcircled{a}\textcircled{+}} & \textasciitilde 15 & 71.26 & 75.70 & 76.13 \\ \bottomrule
   \end{tabular}
        }
\end{center}
\end{table}

\begin{table*}[t]
    \caption{Comparison of ViLT-B/32 with other VLP models on downstream zero-shot retrieval tasks. We exclude the models of which zero-shot retrieval performances were not reported in their original papers. {\textdagger} is pre-trained with a 10M proprietary vision-and-language dataset in addition to the 4M dataset of GCC+SBU. {\textcircled{+}} indicates model trained for a longer 200K pre-training steps.}
    \label{tab:retri_zs}
    \begin{center}
        {\fontsize{8.1pt}{8.1pt}\selectfont
    \begin{tabular}{llrcccccccccccc}
        \toprule
        \multirow{3}{*}{\shortstack[l]{Visual\\Embed}} & \multirow{3}{*}{Model} & \multirow{3}{*}{\shortstack[r]{Time\\(ms)}} & \multicolumn{6}{c}{Zero-Shot Text Retrieval} & \multicolumn{6}{c}{Zero-Shot Image Retrieval} \\
        &  & & \multicolumn{3}{c}{Flickr30k (1K)} & \multicolumn{3}{c}{MSCOCO (5K)} & \multicolumn{3}{c}{Flickr30k (1K)} & \multicolumn{3}{c}{MSCOCO (5K)} \\
        &  & & R@1 & R@5 & R@10 & R@1 & R@5 & R@10 & R@1 & R@5 & R@10 & R@1 & R@5 & R@10 \\ \midrule
        \multirow{4}{*}{Region} & ViLBERT & \textasciitilde 900 & - & - & - & - & - & - & 31.9 & 61.1 & 72.8 & - & - & - \\
        & Unicoder-VL & \textasciitilde 925 & 64.3 & 85.8 & 92.3 & - & - & - & 48.4 & 76.0 & 85.2 & - & - & - \\
      & UNITER-Base & \textasciitilde 900 & 80.7 & 95.7 & 98.0 & - & - & - & 66.2 & 88.4 & 92.9 & - & - & - \\ 
      & ImageBERT\Textsuperscript{\textdagger} & \textasciitilde 925 & 70.7 & 90.2 & 94.0 & 44.0 & 71.2 & 80.4 & 54.3 & 79.6 & 87.5 & 32.3 & 59.0 & 70.2 \\ \midrule
      \multirow{2}{*}{Linear} & ViLT-B/32 & \textasciitilde 15 & 69.7 & 91.0 & 96.0 & 53.4 & 80.7 & 88.8 & 51.3 & 79.9 & 87.9 & 37.3 & 67.4 & 79.0 \\
       & ViLT-B/32\Textsuperscript{\textcircled{+}} & \textasciitilde 15 & 73.2 & 93.6 & 96.5 & 56.5 & 82.6 & 89.6 & 55.0 & 82.5 & 89.8 & 40.4 & 70.0 & 81.1 \\ \bottomrule
   \end{tabular}
        }
\end{center}
\vskip -\baselineskip
\end{table*}

\begin{table*}[t]
    \caption{Comparison of ViLT-B/32 with other models on downstream retrieval tasks. We use SCAN for w/o VLP SOTA results. {\textdagger} additionally used GQA, VQAv2, VG-QA for pre-training. {\textdaggerdbl} additionally used the Open Images dataset. {\textcircled{a}} indicates RandAugment is applied during fine-tuning. {\textcircled{+}} indicates model trained for a longer 200K pre-training steps.}
    \label{tab:retri_fine-tune}
    \begin{center}
        {\fontsize{7.8pt}{7.8pt}\selectfont
        \begin{tabular}{llrcccccccccccc}
        \toprule
        \multirow{3}{*}{\shortstack[l]{Visual\\Embed}} & \multirow{3}{*}{Model} & \multirow{3}{*}{\shortstack[r]{Time\\(ms)}} & \multicolumn{6}{c}{Text Retrieval} & \multicolumn{6}{c}{Image Retrieval} \\
        &  & & \multicolumn{3}{c}{Flickr30k (1K)} & \multicolumn{3}{c}{MSCOCO (5K)} & \multicolumn{3}{c}{Flickr30k (1K)} & \multicolumn{3}{c}{MSCOCO (5K)} \\
        &  & & R@1 & R@5 & R@10 & R@1 & R@5 & R@10 & R@1 & R@5 & R@10 & R@1 & R@5 & R@10 \\ \midrule
      \multirow{6}{*}{Region} & w/o VLP SOTA & \textasciitilde 900 & 67.4 & 90.3 & 95.8 & 50.4 & 82.2 & 90.0 & 48.6 & 77.7 & 85.2 & 38.6 & 69.3 & 80.4 \\
      & ViLBERT-Base & \textasciitilde 920 & - & - & - & - & - & - & 58.2 & 84.9 & 91.5 & - & - & - \\
      & Unicoder-VL & \textasciitilde 925 & 86.2 & 96.3 & 99.0 & 62.3 & 87.1 & 92.8 & 71.5 & 91.2 & 95.2 & 48.4 & 76.7 & 85.9 \\
      & UNITER-Base & \textasciitilde 900 & 85.9 & 97.1 & 98.8 & 64.4 & 87.4 & 93.1 & 72.5 & 92.4 & 96.1 & 50.3 & 78.5 & 87.2 \\ 
      & OSCAR-Base\Textsuperscript{\textdagger} & \textasciitilde 900 & - & - & - & 70.0 & 91.1 & 95.5 & - & - & - & 54.0 & 80.8 & 88.5 \\ 
      & VinVL-Base\Textsuperscript{\textdagger}\Textsuperscript{\textdaggerdbl} & \textasciitilde 650 & - & - & - & 74.6 & 92.6 & 96.3 & - & - & - & 58.1 & 83.2 & 90.1 \\
      \midrule
      \multirow{2}{*}{Grid} & Pixel-BERT-X152 & \textasciitilde 160 & 87.0 & 98.9 & 99.5 & 63.6 & 87.5 & 93.6 & 71.5 & 92.1 & 95.8 & 50.1 & 77.6 & 86.2 \\
      & Pixel-BERT-R50 & \textasciitilde 60 & 75.7 & 94.7 & 97.1 & 59.8 & 85.5 & 91.6 & 53.4 & 80.4 & 88.5 & 41.1 & 69.7 & 80.5 \\
      \midrule
      \multirow{3}{*}{Linear} & ViLT-B/32 & \textasciitilde 15 & 81.4 & 95.6 & 97.6 & 61.8 & 86.2 & 92.6 & 61.9 & 86.8 & 92.8 & 41.3 & 72.0 & 82.5 \\
       & ViLT-B/32\Textsuperscript{\textcircled{a}} & \textasciitilde 15 & 83.7 & 97.2 & 98.1 & 62.9 & 87.1 & 92.7 & 62.2 & 87.6 & 93.2 & 42.6 & 72.8 & 83.4 \\
       & ViLT-B/32\Textsuperscript{\textcircled{a}\textcircled{+}} & \textasciitilde 15 & 83.5 & 96.7 & 98.6 & 61.5 & 86.3 & 92.7 & 64.4 & 88.7 & 93.8 & 42.7 & 72.9 & 83.1 \\ \bottomrule
   \end{tabular}
        }
\end{center}
\end{table*}

We evaluate ViLT-B/32 on two commonly used datasets: VQAv2 and NLVR2.
We use a two-layer MLP of hidden size 1,536 as the fine-tuned downstream head.

\paragraph{Visual Question Answering.}
The VQAv2 task asks for answers given pairs of an image and a question in natural language.
The annotated answers are originally in free-form natural language, but it is a common practice to convert the task to a classification task with 3,129 answer classes.
Following this practice, we fine-tune ViLT-B/32 on the VQAv2 train and validation sets while reserving 1,000 validation images and their related questions for internal validation.

We report the test-dev score results\footnote{VQA score is calculated by comparing the inferred answer to 10 ground-truth answers: see \url{https://visualqa.org/evaluation.html} for details.} from the submission to the evaluation server.
ViLT falls short of VQA score compared to other VLP models with a heavy visual embedder.
We suspect a detached object representation generated by the object detector eases the training of VQA since questions in VQA typically ask about objects.

\paragraph{Natural Language for Visual Reasoning.}
The NLVR2 task is a binary classification task given triplets of two images and a question in natural language.
As there are two input images unlike the pre-training setup, multiple strategies exist\footnote{UNITER proposed three downstream head setups: pair, triplet, and pair-biattn.}.
Following OSCAR \citep{li2020oscar} and VinVL \citep{zhang2021vinvl}, we use the \textit{pair} method.
Here, the triplet input is reformulated into two pairs (question, image1) and (question, image2), and each pair goes through the ViLT.
The head takes the concatenation of two pooled representations ($p$) as input and outputs the binary prediction.

\cref{tab:clas} shows the results.
ViLT-B/32 maintains competitive performance on both datasets considering its remarkable inference speed.

\subsection{Retrieval Tasks}

We fine-tune ViLT-B/32 on the \citet{karpathy2015deep} split of MSCOCO and F30K.
For image-to-text and text-to-image retrieval, we measure both zero-shot and fine-tuned performance\footnote{R@K corresponds to whether the ground truth is included among top K results from the validation set.}.
We initialize the similarity score head from the pre-trained ITM head, particularly the part that computes the true-pair logits.
We sample 15 random texts as negative samples and tune the model with cross-entropy loss that maximizes the scores on positive pairs.

We report the zero shot retrieval results in \cref{tab:retri_zs} and fine-tuned results in \cref{tab:retri_fine-tune}. 
At zero-shot retrieval, ViLT-B/32 performs better in general than ImageBERT despite ImageBERT's pre-training on a larger (14M) dataset.
At fine-tuned retrieval, recalls for ViLT-B/32 are higher by a large margin than the second fastest model (Pixel-BERT-R50).

\subsection{Ablation Study}
\label{sec:ablation}

\begin{table*}[t]
    \caption{Ablation study of ViLT-B/32. \textcircled{w} denotes whether whole word masking is used for pre-training. \textcircled{m} denotes whether MPP objective is used for pre-training. \textcircled{a} denotes whether RandAugment is used during fine-tuning.}
    \label{tab:ablation}
    \begin{center}
        {\fontsize{8pt}{8pt}\selectfont
        % {\small
    \begin{tabular}{ccccccccccc}
        \toprule
        \multirow{2}{*}{\shortstack[*]{Training\\Steps}} & \multicolumn{3}{c}{Ablation} & VQAv2 & \multicolumn{2}{c}{NLVR2} & \multicolumn{2}{c}{Flickr30k R@1 (1K)} & \multicolumn{2}{c}{MSCOCO R@1 (5K)} \\
        & \textcircled{w} & \textcircled{m} & \textcircled{a} & test-dev & dev & test-P & TR (ZS) & IR (ZS) & TR (ZS) & IR (ZS) \\ \midrule
        25K & X & X & X & 68.96 $\pm$ 0.07 & 70.83 $\pm$ 0.19 & 70.83 $\pm$ 0.23 & 75.39 (45.12) & 52.52 (31.80) & 53.72 (31.55) & 34.88 (21.58) \\
        50K & X & X & X & 69.80 $\pm$ 0.01 & 71.93 $\pm$ 0.27 & 72.92 $\pm$ 0.82 & 78.13 (55.57) & 57.36 (40.94) & 57.00 (39.56) & 37.47 (27.51) \\
        100K & X & X & X & 70.16 $\pm$ 0.01 & 73.54 $\pm$ 0.02 & 74.15 $\pm$ 0.27 & 79.39 (66.99) & 60.50 (47.62) & 60.15 (51.25) & 40.45 (34.59) \\
        100K & O & X & X & 70.33 $\pm$ 0.01 & 74.41 $\pm$ 0.21 & 74.57 $\pm$ 0.09 & 81.35 (69.73) & 61.86 (51.28) & 61.79 (53.40) & 41.25 (37.26) \\
        100K & O & O & X & 70.21 $\pm$ 0.05 & 72.76 $\pm$ 0.50 & 73.54 $\pm$ 0.47 & 78.91 (63.67) & 58.76 (46.96) & 59.53 (47.75) & 40.08 (32.28) \\
        \midrule
        100K & O & X & O & 70.85 $\pm$ 0.13 & 74.91 $\pm$ 0.29 & 75.57 $\pm$ 0.61 & 83.69 (69.73) & 62.22 (51.28) & 62.88 (53.40) & 42.62 (37.26) \\
        200K & O & X & O & 71.26 $\pm$ 0.06 & 75.70 $\pm$ 0.32 & 76.13 $\pm$ 0.39 & 83.50 (73.24) & 64.36 (54.96) & 61.49 (56.51) & 42.70 (40.42) \\ \bottomrule
    \end{tabular}
        }
\end{center}
\end{table*}

In \cref{tab:ablation}, we perform various ablations.
More training steps, whole word masking, and image augmentation come to be beneficial, whereas an additional training objective does not help.

It has been reported that the number of training iterations affects the performance of self-supervised models \citep{devlin2019bert,chen2020simple,chen2020improved}.
As VLP is also a form of self-supervised training, we examine the effects of training durations.
As expected, the performance constantly increases as we train the model for longer training steps (rows 1\textasciitilde 3).
Masking whole words for the MLM objective (rows 3\textasciitilde 4) and fine-tuning with augmentation (row 6) also drive performance.
Further increase in training iterations to 200K improved performance on VQAv2, NLVR2, and zero-shot retrieval.
We stop increasing the number of iterations over 200K as the fine-tuned text retrieval performance decreases afterward.

An additional masked region modeling (MRM) objective has been the key for performance boost in VLP models such as \citet{chen2019uniter}.
We experiment with masked patch prediction (MPP) \citep{dosovitskiy2020image} which mimics the effect of MRM in a form compatible with patch projections. 
The patch $v$ is masked with the probability of 0.15, and the model predicts the mean RGB value of the masked patch from its contextualized vector $z^D_{\text{masked}}|_v$.
However, MPP turns out not to be contributing to downstream performance (rows 4\textasciitilde 5). 
This result is in sharp contrast to the MRM objective on supervision signals from object detection.

\subsection{Complexity Analysis of VLP Models}
\label{subsec:runtime}

\begin{table}[t]
    \caption{Comparison of VLP models in terms of parameter size, FLOPs, and inference latency. Since FLOPs are proportional to input size, we denote the number of input tokens (image+text) as superscripts ("?" when text length is unreported; we arbitrarily use length 40). Although not captured in FLOPs count nor parameter size (because it is not a tensor operation), note that \textit{per-class} NMS for 1,600 classes amounts to more than 500 ms in latency. NMS latency varies a lot according to the number of detected classes.}
    \label{tab:runtime}
    \begin{center}
        {\fontsize{8.4pt}{8.4pt}\selectfont
    \begin{tabular}{llrrr}
        \toprule
        \multirow{2}{*}{\shortstack[l]{Visual\\Embed}} & \multirow{2}{*}{Model} &  \multirow{2}{*}{\shortstack[r]{\#Params\\(M)}} & \multirow{2}{*}{\shortstack[r]{\#FLOPs\\(G)}} & \multirow{2}{*}{\shortstack[r]{Time\\(ms)}} \\
   & & & & \\ \midrule
   \multirow{8}{*}{Region} & ViLBERT\Textsuperscript{36+36} & 274.3 & 958.1 & \textasciitilde 900 \\
   & VisualBERT\Textsuperscript{36+128} & 170.3 & 425.0 & \textasciitilde 925 \\
   & LXMERT\Textsuperscript{36+20} & 239.8 & 952.0 & \textasciitilde 900 \\
   & UNITER-Base\Textsuperscript{36+60} & 154.7 & 949.9 & \textasciitilde 900 \\
   & OSCAR-Base\Textsuperscript{50+35} & 154.7 & 956.4 & \textasciitilde 900 \\
   & VinVL-Base\Textsuperscript{50+35} & 157.3 & 1023.3 & \textasciitilde 650 \\
   & Unicoder-VL\Textsuperscript{100+?} & 170.3 & 419.7 & \textasciitilde 925 \\
   & ImageBERT\Textsuperscript{100+44} & 170.3 & 420.6 & \textasciitilde 925 \\
   \midrule
   \multirow{2}{*}{Grid} & Pixel-BERT-X152\Textsuperscript{146+?} & 144.3 & 185.8 & \textasciitilde 160 \\
   & Pixel-BERT-R50\Textsuperscript{260+?} & 94.9 & 136.8 & \textasciitilde 60 \\
    \midrule
   Linear & ViLT-B/32\Textsuperscript{200+40} & 87.4 & 55.9 & \textasciitilde 15 \\
   \bottomrule
   \end{tabular}
        }
\end{center}
\end{table}

\begin{table}[t]
    \caption{VLP model components. "PC" is for \textit{per-class} manner NMS and "CA" is for \textit{class-agnostic}. Following \citet{tan2019lxmert}, one single-modality layer is counted as 0.5 multi-modality layer.}
    \label{tab:vlp_components}
    \begin{center}
        {\fontsize{7.4pt}{7.4pt}\selectfont
    \begin{tabular}{lllllr}
        \toprule
        \multirow{2}{*}{\shortstack[l]{Visual\\Embed}} & \multirow{2}{*}{Model} &  \multirow{2}{*}{\shortstack[l]{CNN\\Backbone}} & \multirow{2}{*}{\shortstack[l]{RoI\\Head}} & \multirow{2}{*}{NMS} & \multirow{2}{*}{\shortstack[r]{Trans.\\Layers}} \\
   & & & & \\ \midrule
   \multirow{8}{*}{Region} & ViLBERT & R101 & C4 & PC & \textasciitilde 15 \\
   & VisualBERT & X152 & FPN & PC & 12 \\
   & LXMERT & R101 & C4 & PC & \textasciitilde 12 \\
   & UNITER-Base & R101 & C4 & PC & 12 \\
   & OSCAR-Base & R101 & C4 & PC & 12 \\
   & VinVL-Base & X152 & C4 & CA & 12 \\
   & Unicoder-VL & X152 & FPN & PC & 12 \\
   & ImageBERT & X152 & FPN & PC & 12 \\
   \midrule
   \multirow{2}{*}{Grid} & Pixel-BERT-X152 & X152 & - & - & 12 \\
   & Pixel-BERT-R50 & R50 & - & - & 12 \\
    \midrule
   Linear & ViLT-B/32 & - & - & - & 12 \\
   \bottomrule
   \end{tabular}
        }
\end{center}
\end{table}

\begin{figure*}[t]
    \centering
    \includegraphics[width=1\textwidth]{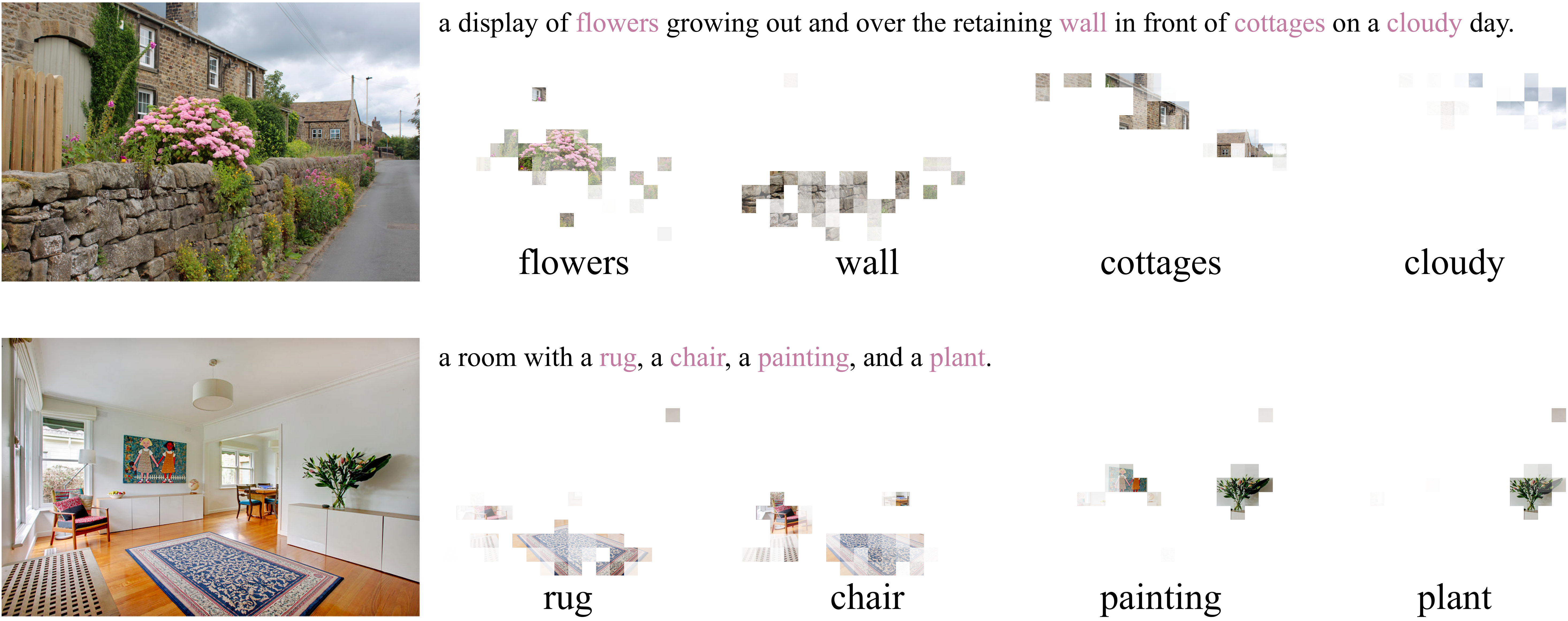}
    \caption{Visualizations of transportation plan of word patch alignment. Best viewed zoomed in.}
    \label{fig:vis}
\end{figure*}

We analyze the complexity of VLP models in various terms.
In \cref{tab:runtime}, we report the number of parameters, the number of floating-point operations (FLOPs), and the inference latency of the visual embedder and transformer.
We exclude the textual embedder because it is shared by all VLP models\footnote{FLOPs and time are neglectable because the operation is an embedding lookup. The 30K embedding dictionary used by \texttt{bert-base-uncased} has 23.47 M parameters}.
The latency is averaged over 10K times on a Xeon E5-2650 CPU and an NVIDIA P40 GPU.

The input size in terms of image resolution and the length of concatenated multimodal input sequence affects the number of FLOPs. We co-note the sequence lengths.
The image resolution is 800 $\times$ 1,333 for region-based VLP models and Pixel-BERT-R50, 600 $\times$ 1,000 for Pixel-BERT-X152, and 384 $\times$ 640 for ViLT-B/32.

In Pixel-BERT and ViLT, visual tokens are sampled during pre-training and used in full during fine-tuning. We report the maximum number of visual tokens. 

We observe that the runtime of BERT-base-like transformers varies only by $<1$ ms for input sequences of length under 300. Since patch projection of ViLT-B/32 generates at most 240 image tokens, our model can still be efficient even though it receives a combination of image and text tokens.

\subsection{Visualization}

\cref{fig:vis} is an example of a cross-modal alignment.
The transportation plan of WPA expresses a heatmap for a text token highlighted in pink color. 
Each square tile represents a patch, and its opacity indicates how much mass is transported from the highlighted word token.

More IPOT iterations-- more than over 50 as in the training phase-- help the visualization heatmap converge; empirically, 1,000 iterations are sufficient to get a clearly identifiable heatmap.
We z-normalize the plan for each token and clamp the values to [1.0, 3.0].

\section{Conclusion and Future Work}
\label{sec:conclusion}
In this paper, we present a minimal VLP architecture, Vision-and-Langauge Transformer (ViLT).
ViLT is competent to competitors which are heavily equipped with convolutional visual embedding networks (e.g., Faster R-CNN and ResNets).
We ask for future work on VLP to focus more on the modality interactions inside the transformer module rather than engaging in an arms race that merely powers up unimodal embedders.

Although remarkable as it is, ViLT-B/32 is more of a proof of concept that efficient VLP models free of convolution and region supervision can still be competent.
We wrap up by pointing out a few factors that may add to the ViLT family.

\paragraph{Scalability.}
As shown in papers on large-scale transformers \citep{devlin2019bert,dosovitskiy2020image}, the performance of pre-trained transformers scale well given an appropriate amount of data. 
This observation paves the way for even better performing ViLT variants (e.g., ViLT-L (large) and ViLT-H (huge)).
We leave training larger models for future work because aligned vision-and-language datasets are yet scarce. 

\paragraph{Masked Modeling for Visual Inputs.}
Considering the success of MRM, we speculate that the masked modeling objective for the visual modality helps by preserving the information up until the last layer of the transformer. 
However, as observed in \cref{tab:ablation}, a naive variant of MRM on image patches (MPP) fails.

\citet{cho2020x} proposed to train their grid RoIs on masked object classification (MOC) tasks.
However, the visual vocabulary cluster in this work was fixed during the vision and language pre-training together with the visual backbone. For trainable visual embedders, one-time clustering is not a viable option.
We believe that alternating clustering \citep{caron2018deep,caron2019unsupervised} or simultaneous clustering \citep{asano2019self,caron2020unsupervised} methods studied in visual unsupervised learning research could be applied.

We encourage future work that does not use region supervision to devise a more sophisticated masking objective for the visual modality.

\paragraph{Augmentation Strategies.}
Previous work on contrastive visual representation learning \citep{chen2020simple, chen2020improved} showed that gaussian blur, not employed by RandAugment, brings noticeable gains to downstream performance compared with a simpler augmentation strategy \citep{he2020momentum}.
Exploration of appropriate augmentation strategies for textual and visual inputs would be a valuable addition.

\clearpage

\bibliography{vilt}
\bibliographystyle{icml2021}

\end{document}